\documentclass{article}

\usepackage{booktabs} % For professional table formatting
\usepackage{multirow} % For multirow and multicolumn headers

\usepackage[final,nonatbib]{neurips_2023}
\PassOptionsToPackage{numbers,compress}{natbib}
\usepackage{natbib}

\usepackage[utf8]{inputenc} % allow utf-8 input
\usepackage[T1]{fontenc}    % use 8-bit T1 fonts
\usepackage{hyperref}       % hyperlinks
\usepackage{url}            % simple URL typesetting
\usepackage{booktabs}       % professional-quality tables
\usepackage{amsfonts}       % blackboard math symbols
\usepackage{nicefrac}       % compact symbols for 1/2, etc.
\usepackage{microtype}      % microtypography
\usepackage{xcolor}         % colors
\usepackage{graphicx} % For including images
\usepackage{float}    % To control float positioning
\usepackage{amsmath}

\title{Differential MultiModal Transformers}

\author{%
  Jerry Li \\
  University of California, Riverside \\
  \texttt{jli793@ucr.edu} \\
  \And
    Timothy Oh \\
  University of California, Riverside \\
  \texttt{toh012@ucr.edu} \\
  \And
    Joseph Hoang \\
  University of California, Riverside \\
  \texttt{jhoan075@ucr.edu} \\
\And
  Vardhit Veeramachaneni \\
  University of California, Riverside \\
  \texttt{vveer003@ucr.edu} \\
}

\begin{document}

\maketitle

\begin{abstract}
    Small language models have gained significant popularity due to their efficiency and growing capabilities. However, incorporating additional modalities, such as vision, can exacerbate the challenge of limited context windows by introducing noise. Recent studies have highlighted that Transformer attention mechanisms often disproportionately focus on irrelevant contexts. In this work, we extend the Differential Attention mechanism, originally designed for text-only models, to the text-vision model PaliGemma. Our aim is to evaluate its ability to mitigate noisy information retrieval and reduce hallucinations. To this end, we fine-tuned the PaliGemma 3B model using LoRA, incorporating Differential Attention, and experimented with various parameter settings and configurations. We demonstrate that Differential Attention can be adapted and integrated into the fine-tuning of existing models to enhance noisy information retrieval and question-answering capabilities.
\end{abstract}

\section{Introduction}

Large Language Models in the recent years have seen extreme growth in efficiency and multi-modality. Models such as \cite{gpt4omini} \cite{phi3}\cite{smallgemini} have demonstrated exceptional performance for their rather compact size when compare to larger counterparts. However there have been more recent studies on the attention mechanism behind these models that shown attention is still not fully utilized \cite{differentialtransformer} \cite{softmaxnotgood}. One work, Differential Transformers \cite{differentialtransformer} uncovers how vanilla attention is prone to noise from text inputs. To combat this they propose a new type attention, Differential Attention, which demonstrates hallucination and information retrieval improvements while maintaining efficiency. 

However the avenue of multi-modality of this attention mechanism remains unexplored. Multimodal Large Language Models (MLLMs) have demonstrated an impressive understanding and integrating modalities beyond text, such as images, audio, and video. This allows for more comprehensive reasoning and contextual understanding, enabling applications like image captioning, visual question answering (VQA), and cross-modal retrieval \cite{docvqa}. Furthermore these MLLMs are becoming more efficient and compact with smaller weights while still providing multimodal capabilities. At the same time, more modalities can introduce further noise that can district the model in tasks such as information retrieval \cite{liu2023lostmiddlelanguagemodels} \cite{wang2024comprehensivereviewmultimodallarge}.

PaliGemma \cite{paligemma} is an open-source 3B parameter text-image multimodal model from Google that combines the Siglip vision encoder and Gemma text decoder. The Siglip encoder \cite{siglip} processes images into embeddings using a vision transformer (ViT) which are then projected into a shared space with text tokens via a multimodal projector. The Gemma text decoder takes these fused tokens as input to generate text outputs with image captioning, visual question answering (VQA), and multimodal reasoning capabilities. This architecture allows PaliGemma to achieve efficient cross-modal reasoning with a relatively small parameter count, however at a cost of a larger textual context length. 

Given these challenges we analyze how differential attention can be used to mitigate attention noisy in multimodal models through integrating it into PaliGemma. We propose finetuning the pretrained PaliGemma weights with LoRA while also training newly added parameters required for the Differential Attention. The dataset we use is VQAv2 \cite{vqav2}, a multimodal text-image dataset with questions that span multiple categories such as object recognition, counting, and reasoning. In order to demonstrate any possible improvements with Differential Attention, we utilize the Multimodal Needle in Haystack evaluation \cite{needlemultimodalhaystack}, a benchmark for assessing the retrieval capabilities of multimodal models when given a pool of distractors. Overall, we introduce Differential PaliGemma, a fine-tuned version of PaliGemma incorporating Differential Attention, which demonstrates enhanced performance in noisy information retrieval and question answering tasks. Our code can be found [\href{https://github.com/Jeli04/CS228-Project}{here}].

\section{Background and Related Works}

\subsection{Multimodal Language Models}

MLLMs also extend the capability of unimodal models by involving diverse modalities, including images, audio, and videos, into their reasoning frameworks. In this regard, these models endeavor to integrate information from many modalities to enable several tasks such as VQA, cross-modal retrieval, and image captioning. What characterizes these models best is their ability to contextualize information across modalities, achieving more comprehensive understandings and generating richer outputs compared to unimodal models.

Recent breakthroughs like CLIP \cite{clip} and Flamingo \cite{flamingo} have indeed established that a single latent space where representations from both vision and text come together confers unprecedented generalization powers. Nonetheless, there are challenges while dealing with MLLMs; in particular, balancing the representational power of each modality, avoiding overfitting to one modality, and mitigating the increased noise due to the larger context window. These issues are particularly critical in tasks that require fine-grained reasoning, since irrelevant features from one modality can overshadow relevant ones from another.

As MLLMs become increasingly complex, there is increasing focus on the development of efficient, compact architectures that will be able to realize multimodal reasoning without excessive computational overhead. Models such as PaLI \cite{pali}, phi-3.5-Vision \cite{phi3}, PaliGemma \cite{paligemma} illustrate the trend toward smaller, more efficient designs, integrating vision and language with carefully optimized architectures. However, sustaining performance under these constraints requires innovative attention mechanisms or fine-tuning strategies, as shown in this work.

\subsection{LoRA Finetuning}
LoRA \cite{lora} is an effective fine-tuning approach that dramatically saves computational cost and memory requirements towards adapting pre-trained models into specific tasks. Normal fine-tuning updates the whole parameters of the model. Therein, this is prohibitive for large-scale models, such as transformers, by its very large number of parameters. LoRA solves it by freezing the pre-trained weights of the model but incorporates learnable low-rank matrices into the forward pass of specific layers.

In practice, LoRA injects trainable rank-decomposition matrices into the attention mechanism of transformers, modifying only a small subset of the model's parameters during fine-tuning. For an attention layer with weights 
$W_q \in \mathbb{R}^{d_{\text{model}} \times d_k}$, LoRA adds a low-rank update:

\begin{equation}
\begin{aligned}
    W_q' = W_q + \Delta W_q, \quad \Delta W_q = AB
\end{aligned}
\label{eq:lora}
\end{equation}

where $A \in \mathbb{R}^{d_{\text{model}} \times r}$ and $B \in \mathbb{R}^{r \times d_k}$, with $r \ll d_{\text{model}}, d_k$. The pretrained weights $W_q$ remain frozen, and only $A$ and $B$ are optimized. This approach retains the original capability of the model while offering task-specific adaptability due to minimal adjustments of parameters.

LoRA has been especially effective in scenarios with either limited computational resources or a scarcity of training data. It allows models, such as PaliGemma, to be fine-tuned efficiently on task-specific layers for multimodal tasks. In this work, we leverage LoRA to adapt PaliGemma to include the Differential Attention mechanism, enabling the evaluation of its impact on multimodal reasoning without having to retrain the full model from scratch.

\section{Differential PaliGemma}

\begin{figure}[H]  
    \centering
    \includegraphics[width=0.8\linewidth]{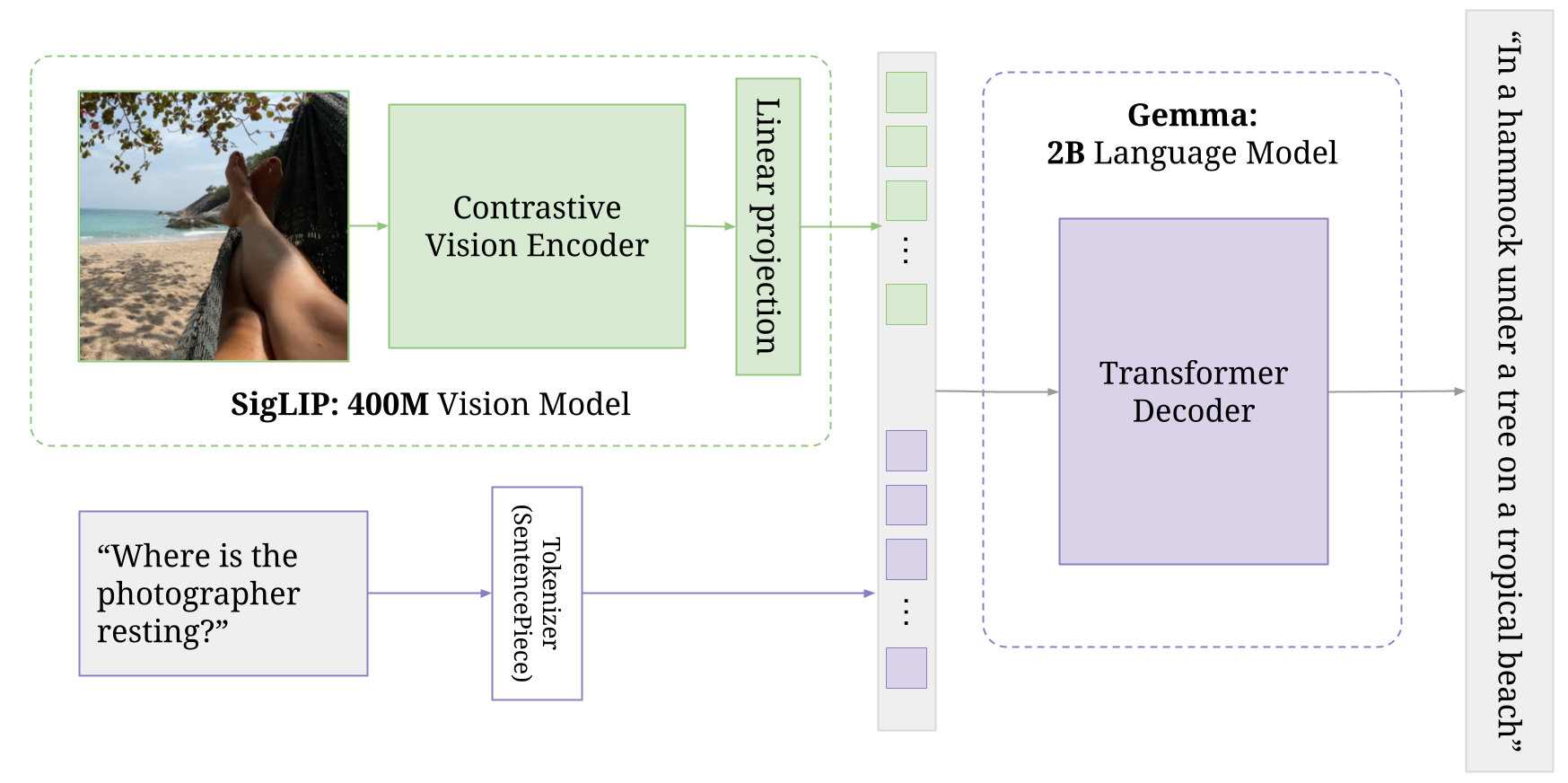}
    \caption{PaliGemma architecture.}
    \label{fig:paligemma} % Use \ref{fig:example_image} in the text to reference this figure
\end{figure}

\subsection{Architecture Overview}
The PaliGemma architecture shown in Figure \ref{fig:paligemma} consists of two parts: the image encoder and text decoder. The image decoder is Siglip, the contrastive model built ontop of CLIP \cite{clip}. Siglip is a ViT that learns the image spatial features and outputs it through a linear projection to a unified space with the text tokens. The raw text is tokenized through the SentencePiece \cite{sentencepiece} tokenizer and than concatenated with the image tokens from the linear projection. This than is fed to the text decoder, Gemma 2B \cite{gemma}, which follows the core structure of a decoder-only Transformer used in LLMs. 

\subsection{Original Differential Attention}
We propose to modify the self attention layers in both Siglip and Gemma to use Differential Attention instead. The core idea of Differential Attention is to create two sets of queries and keys that undergo self attention and than differentiate them as shown below. The input $X \in \mathbb{R}^{N\times d_{model}}$, and $Q_1, Q_2, K_1, K_2 \in \mathbb{R}^{N \times d}$ and $V \in \mathbb{R}^{N \times 2d}$, where $d = d_{head} / 2$.

\begin{equation}
\begin{aligned}
    [Q_1; Q_2] &= XW^Q, \quad [K_1; K_2] = XW^K, \quad V = XW^V, \\
    \text{DiffAttn}(X) &= \left(\text{softmax}\left(\frac{Q_1 K_1^T}{\sqrt{d}}\right) - \lambda \, \text{softmax}\left(\frac{Q_2 K_2^T}{\sqrt{d}}\right)\right)V
\end{aligned}
\label{eq:diff-attention}
\end{equation}

Where the weights $W^Q, W^K, W^V \in \mathbb{R}^{d_{\text{model}} \times 2d}$. $\lambda$ is created from learnable parameters for each query and key in both sets.

\begin{equation}
\lambda = \exp(\lambda_{q_1} \cdot \lambda_{k_1}) - \exp(\lambda_{q_2} \cdot \lambda_{k_2}) + \lambda_{\text{init}}
\label{eq:lambda}
\end{equation}

where $\lambda_{q_1}, \lambda_{k_1}, \lambda_{q_2}, \lambda_{k_2} \in \mathbb{R}^d$ are learnable vectors, and $\lambda_{\text{init}} \in (0, 1)$ is $\lambda$'s initialization constant. As in the paper, setting

\begin{equation}
\lambda_{\text{init}} = 0.8 - 0.6 \times \exp(-0.3 \cdot (l - 1))
\label{eq:lambda_init}
\end{equation}

works well in practice, where $l \in [1, L]$ represents the layer index. 

\subsection{Differential Attention in PaliGemma}

Since we are finetuning PaliGemma with additional parameters, we must preserve as much stability in the original attention weights as possible while also being able to reduce the attention noise. Rather than creating two separate sets of queries and keys like the original Differential Attention, we take the original query and keys and simply duplicate itself as shown below. Since we are using a single set of queries and keys, the input $X \in \mathbb{R}^{N\times d_{model}}$, and $Q, K \in \mathbb{R}^{N \times d_{head}}$ and $V \in \mathbb{R}^{N \times d_{head}}$.

\begin{equation}
\begin{aligned}
    [Q] &= XW^Q, \quad [K] = XW^K, \quad V = XW^V, \\
    \text{DiffAttn}(X) &= \left(\text{softmax}\left(\frac{Q K^T}{\sqrt{d_{\text{head}}}}\right) - \lambda \, \text{softmax}\left(\frac{Q K^T}{\sqrt{d_{\text{head}}}}\right)\right)V
    \label{eq:diff-attention}
\end{aligned}    
\end{equation}

Where the weights $W^Q, W^K, W^V \in \mathbb{R}^{d_{\text{model}} \times d_{\text{head}}}$. Equations \ref{eq:lambda} and \ref{eq:lambda_init} remain the same except $\lambda_{q_1}, \lambda_{k_1}, \lambda_{q_2}, \lambda_{k_2} \in \mathbb{R}^{d_{\text{head}}}$ instead. 

In theory, since we are using pretrained attention weights, there may be no need for a second set of distinct queries and keys. Instead, the concept of subtracting two sets of attention weights to reduce noise can be directly applied to a single predefined set. As the parameter $\lambda$ scales the influence of the secondary attention weights, the model, during fine-tuning, should be able to learn optimal $\lambda$ values. This allows the model to diminish attention on noisy or irrelevant information while enhancing focus on more critical and meaningful areas of the input.

\textbf{Multi-Head Differential Attention} As in the Differential Transformer’s paper, we adopt the multi-head mechanism in our architecture. Let $h$ denote the number of attention heads. For each head $i \in [1, h]$, we utilize distinct projection matrices $W^Q_i, W^K_i, W^V_i$. A scalar $\lambda$ is shared across heads within the same layer. The head outputs are normalized and combined as follows:

\begin{equation}
\begin{aligned}
    \text{head}_i &= \text{DiffAttn}(X; W^Q_i, W^K_i, W^V_i, \lambda) \\
    \overline{\text{head}}_i &= (1 - \lambda_{\text{init}}) \cdot \text{LN}(\text{head}_i) \\
    \text{MultiHead}(X) &= \text{Concat}(\overline{\text{head}_1}, \ldots, \overline{\text{head}_h}) W^O 
\end{aligned} 
\label{eq:multi-head-diff}
\end{equation}

Here, $\lambda_{\text{init}}$ is a constant scalar initialized as in Equation \ref{eq:lambda_init}. $W^O \in \mathbb{R}^{d_{\text{model}} \times d_{\text{model}}}$ is a learnable projection matrix, and $\text{LN}(\cdot)$ represents RMSNorm. The concatenation of head outputs occurs along the channel dimension. To align gradients with the Transformer, we scale $\text{LN}(\cdot)$ by $(1 - \lambda_{\text{init}})$. According to Appendix F from \cite{differentialtransformer}, this design maintains stable gradient flow and ensures efficient training with similar hyperparameters. We also set the number of heads to $h = d_{\text{model}} / d_{\text{head}}$ which is the standard. 

\textbf{Differential PaliGemma Layer} Also as in the Differential Transformer’s paper, the overall architecture stacks $L$ layers, where each layer contains a multi-head differential attention module and a feed-forward network module. We describe the Differential Transformer layer as:

\begin{equation}
\begin{aligned}
    Y^l &= \text{MultiHead}(\text{LN}(X^l)) + X^l \\
    X^{l+1} &= \text{SwiGLU}(\text{LN}(Y^l)) + Y^l 
\end{aligned}
\label{eq:multi-head-diff}
\end{equation}

where $\text{LN}(\cdot)$ is RMSNorm, $\text{SwiGLU}(X) = (\text{swish}(XW^G) \odot XW_1)W_2$, and $W^G, W_1 \in \mathbb{R}^{d_{\text{model}} \times \frac{8}{3} d_{\text{model}}}, W_2 \in \mathbb{R}^{\frac{8}{3} d_{\text{model}} \times d_{\text{model}}}$ are learnable matrices. In our finetuning process we experiment with both the SwiGLU from Equation \ref{eq:multi-head-diff} and the original PaliGemma MLP to see which one performs better. We also compare the modified Differential Attention for finetuning that we proposed with the original method.

% \textbf{Applying Modifications to Paligemma} In the original attention paper \cite{attentionisallyouneed}, which is what Siglip and the Gemma architectures both use, the attention layer takes in an input $X \in \mathbb{R}^{N \times d_{model}}$ and gives an output $Y \in \mathbb{R}^{N \times d_{model}}$, which is the same size. Since the differential attention layer outputs the exact same dimensions, it can readily replace the original Attention layer that Siglip and Gemma use. This is what we have done for our paper.

\section{Finetuning}

We conduct multiple finetuning runs on the original PaliGemma and also variations of the Differential PaliGemma. In order to determine the best method of Differential Attention we tested variations of it, as shown in Table \ref{table:model-setup}. For consistency we only compared results with the same finetuning hyperparamters across the different variations and model types. 

\subsection{Dataset}
The dataset utilized for fine-tuning and evaluation was VQAv2 \cite{vqav2}, a widely adopted benchmark in the visual question answering (VQA) domain. This dataset contains over 440,000 image-question pairs and serves as a standard for evaluating the reasoning capacity of multimodal models. To ensure the robustness of the evaluation, we adhered to the official VQAv2 scoring method, which goes beyond simple exact word matching. Instead, VQAv2 employs a majority-voting system to account for human-level variability in responses.

For each question, 10 annotators provide independent answers. The model’s predicted answer is compared to these human-provided responses, and the evaluation score is computed using the following formula:

\begin{equation}
    \text{Score}(a) = \min\left( \frac{n(a)}{3}, 1 \right)
    \label{eq:vqav2_score}
\end{equation}

Where $n(a)$ represents the number of annotators who responed with the same answer as the model. These human-provided responses come alongside the dataset which we downloaded through Huggingface. 

\begin{table}[h]
  \caption{Model Setup and VQAv2 Scores}
  \label{table:model-setup}
  \centering
  \begin{tabular}{lllllll} % Align all columns properly
    \toprule
    Model & Diff Attention & Feedforward  & Learning Rate & LoRA Rank/Alpha & Weight Decay & VQAv2 Score \\
    \midrule
    1 & No            & MLP           & $4 \times 10^{-4}$ & 32/64 & 1e-9 & \textbf{53.21}\\
    2 & Yes           & SwiGLU        & $4 \times 10^{-4}$ & 32/64 & 1e-9 & \textbf{51.91} \\
    3 & No            & MLP           & $2 \times 10^{-5}$ & 16/32 & 1e-8 & 35.72\\
    4 & Yes           & MLP           & $2 \times 10^{-5}$ & 32/64 & 1e-8 & 33.91\\
    5 & Yes (Original) & SwiGLU       & $2 \times 10^{-5}$ & 32/64 & 1e-8 & 41.32\\
    \bottomrule
  \end{tabular}
  \centering
\end{table}

\subsubsection{Finetuning Process}
To identify the most effective model configuration, we employed a multi-stage fine-tuning and evaluation strategy. Each model was fine-tuned using the Adam optimizer with a batch size of 4. For picking the learning rate we also tried following the $\text{constant} = \frac{\text{lr}}{\text{batch size}}$
 where the constant we used was $1e-4$. We also followed the standard relationship between the LoRA rank and alpha of $alpha = rank \times 2$ Given the constraints of time, computational resources, and early empirical observations, we didn't experiment with every combination of model setup.

Our approach consisted of two key parts. \textbf{Part 1} involved fine-tuning multiple candidate models, including versions with and without Differential Attention, on a subset comprising 40\% of the VQAv2 training dataset. This allowed us to determine which model setup to use for the Multimodal Needle in Haystack Benchmark while saving time. From this stage, we selected the two best-performing models: one incorporating Differential Attention and the corresponding one without. \textbf{Part 2} involved fine-tuning both of these two models on the full VQAv2 training set for one epoch.

Based on the VQAv2 results from \ref{table:model-setup}, we determined that model 1 and 2 had the best potential based on their scores. We would use these two models in the follow section on the Multimodal Needle in Haystack Benchmark for evaluations.

\section{Evaluations}

To comprehensively assess the performance of the differential attention mechanism in reducing noisy information and mitigating hallucinations, we employ the MultiModal Needle-in-a-Haystack (MMNeedle) benchmark. This benchmark evaluates two critical capabilities of the model in a single inference pass: (1) understanding the semantics of both visual and textual inputs, and (2) locating target images (needles) within long-context inputs (haystack) \cite{needlemultimodalhaystack}. 

The MMNeedle benchmark presents models with a haystack of images, where each image is divided into an \( N \times N \) grid of sub-images. Additionally, the model receives a text caption describing the sub-image (needle) to be identified. The task requires the model to identify the image containing the needle and pinpoint its exact row and column indices.

For this evaluation, we leverage the MS COCO 2014 validation dataset as the source for constructing the MMNeedle dataset.

\subsection{Needle in the Haystack: Experimental Setup}

Due to the Paligemma model's inherent constraints, including its relatively small size (3 billion parameters) and a limited input token capacity (\(\sim 180\) tokens), the evaluation process was adapted accordingly. The experimental procedure is as follows:

\begin{enumerate}
    \item \textbf{Preprocessing Input}: All input images are resized to a resolution of \( 224 \times 224 \) pixels and converted to the RGB format to ensure compatibility with the model's input expectations.

    \item \textbf{Stitching Images}: An \( N \times N \) grid is constructed by stitching sub-images together. For our evaluation, we set \( N = 2 \), resulting in each stitched image containing four sub-images.

    \item \textbf{Selecting the Needle}: One sub-image within the stitched grid is selected as the needle, along with its corresponding textual caption from the dataset.

    \item \textbf{Prompt Construction}: Given the model's input token limit, a concise prompt is constructed in the format:
    \begin{quote}
        \texttt{Caption + "Where is the caption? Top or Bottom?"}
    \end{quote}
    This simplification allows the model to process the input efficiently.

    \item \textbf{Iterative Querying}: To identify the needle's location, the model is queried in two steps:
    \begin{enumerate}
        \item Determine whether the needle is located in the top or bottom half of the grid.
        \item Determine whether the needle is on the left or right side.
    \end{enumerate}

    \item \textbf{Mapping Coordinates}: Based on the model's responses, the outputs are mapped to a coordinate index system to derive numerical results.

    \item \textbf{Sample Limitation}: Due to computational constraints and hardware limitations, we evaluate the model's performance on a subset of 200 samples from the constructed dataset.
\end{enumerate}

\subsection{Evaluation Results}

\begin{figure}[H]  
    \centering
        \includegraphics[width=.9\linewidth]{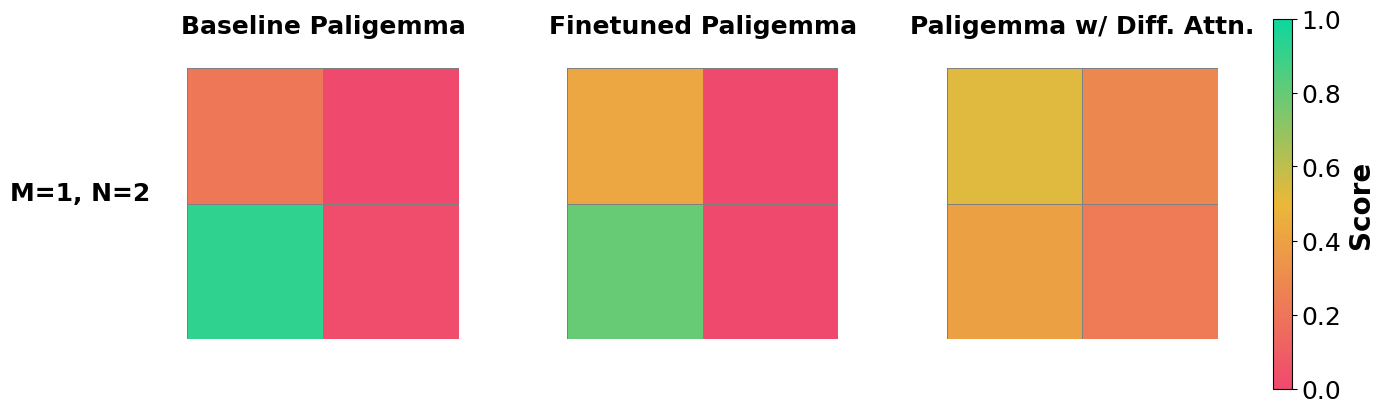} 
    \caption{MMNeedle evaluation performance comparison. The x-axis plots each model's performance, while the Y-axis shows the evaluation setup. M = 1 represents one stitched input image, and N = 2 represents a 2x2 stitched image containing 4 subplot images. Each cell represents the accuracy in which the model correctly predicts that the needle is within that cell. A \textcolor{red}{redder} cell indicates lower accuracy, while a \textcolor{green}{greener} cell indicates higher accuracy.}
    \label{fig:results_10_heatmap} 
\end{figure}

\begin{table}[h!]
\centering
\begin{tabular}{lc}
\toprule
\multicolumn{1}{l}{\textbf{Stitching}} & \multicolumn{1}{c}{\textbf{2 $\times$ 2}} \\ % Adjusted row
\cmidrule(r){2-2}
\textbf{Metrics} & \textbf{Index Accuracy} \\ % Sub-header
\midrule
Baseline PaliGemma & 28.75 \\
Finetuned PaliGemma & 30.42 \\
Finetuned with Diff. Attn. PaliGemma & \textbf{34.72} \\ % Bold text for emphasis
\bottomrule
\end{tabular}
\vspace{10pt} 
\caption{Accuracy for M=1, N=2 setting. The index accuracy represents the proportion of of samples in which the model correctly predicts the location of the subimage within the stitched image}
\label{table:index_acc}

\end{table}

In this section, we present the results of the MMNeedle benchmark across three configurations of the Paligemma model: the baseline, the baseline finetuned, and the finetuned model incorporating the differential attention mechanism. Figure \ref{fig:results_10_heatmap} compares the performance of these configurations in terms of their ability to accurately predict the location of the sub-image within the stitched image.

For both the baseline and finetuned models, we observe that when the needle is located in the bottom-left of the stitched image, the models achieve high accuracy in predicting its position. However, performance deteriorates when the needle is positioned in the top-left, and significantly worsens when the needle is placed on the right side, highlighting the model's difficulty in attending to the right side of the image. Table \ref{table:index_acc} shows that the index accuracies for the baseline and finetuned models are 28.75\% and 30.42\%, respectively.

When the differential attention mechanism is applied, we notice a decrease in performance for predicting the needle’s position in the bottom-left. Despite this, the model demonstrates improved robustness, showing a better ability to correctly predict the location of the needle across various positions within the stitched image. The inclusion of the differential attention mechanism leads to a notable increase in index accuracy, with this configuration achieving 34.72\% \ref{table:index_acc}.

% \subsection{Tables}

% All tables must be centered, neat, clean and legible.  The table number and
% title always appear before the table.  See Table~\ref{sample-table}.

% Place one line space before the table title, one line space after the
% table title, and one line space after the table. The table title must
% be lower case (except for first word and proper nouns); tables are
% numbered consecutively.

% Note that publication-quality tables \emph{do not contain vertical rules.} We
% strongly suggest the use of the \verb+booktabs+ package, which allows for
% typesetting high-quality, professional tables:
% \begin{center}
%   \url{https://www.ctan.org/pkg/booktabs}
% \end{center}
% This package was used to typeset Table~\ref{sample-table}.

% \begin{table}
%   \caption{Sample table title}
%   \label{sample-table}
%   \centering
%   \begin{tabular}{lll}
%     \toprule
%     \multicolumn{2}{c}{Part}                   \\
%     \cmidrule(r){1-2}
%     Name     & Description     & Size ($\mu$m) \\
%     \midrule
%     Dendrite & Input terminal  & $\sim$100     \\
%     Axon     & Output terminal & $\sim$10      \\
%     Soma     & Cell body       & up to $10^6$  \\
%     \bottomrule
%   \end{tabular}
% \end{table}

\section{Conclusion and Future Directions}

Although Multimodal Large Language Models (MLLMs) exhibit exceptional capabilities in question answering, reasoning, and understanding multiple modalities, these strengths can come with the drawback of increased susceptibility to noise. With this we introduced Differential PaliGemma, a finetuned PaliGemma on a modified version of Differential Attention that demonstrates increased capabilities in noisy information retrieval. Due to time and hardware constraints we were unable to finetune on further epochs and hyperparameter settings, and also conduct larger and deeper evaluations with the Multimodal Needle in the Haystack benchmark due to PaliGemma limitations. In the future we'd like to expand on all of this while also testing other models such as Phi-3.5-Vision as PaliGemma has a limited context window.

\section{Contributions}

Timothy and Jerry experimented with the Differential Attention and finetuning while also conducting the VQAv2 evals to determine which parametres and model confirguation to use. 

Joseph and Vardhit developed the evaluation pipeline, which involved adapting and creating evaluation scripts \cite{needlemultimodalhaystack}. This work also included generating visual representations of the evaluation results and analyzing these outcomes to provide feedback for the fine-tuning process.

\section{GPT Usage}
ChatGPT was used to help learn information while also double checking some of our work. In the code it was only used to help write the structure of VQAv2 eval code. Outside of GPT we used Huggingface resources as well as the official code from the Transformers library as reference. We also used the [\href{https://www.youtube.com/watch?v=vAmKB7iPkWw}{video}] and code as a starting point but added our own modifications and integrations for Differential Attention.

For the displaying the evaluation results, ChatGPT aided in plotting our results into the displayed heatmaps to our preferences. It also aided in the general refinement of the scripts such as bug fixing. 

%\printbibliography

\bibliographystyle{plainnat}   % or whatever style NeurIPS wants
\bibliography{references}      % references.bib still needed locally

%%%%%%%%%%%%%%%%%%%%%%%%%%%%%%%%%%%%%%%%%%%%%%%%%%%%%%%%%%%%

\end{document}